\documentclass[letterpaper, 10 pt, conference]{ieeeconf}  

\IEEEoverridecommandlockouts                              

\usepackage{graphics} 
\usepackage{amsmath} 
\usepackage{amssymb}  
\usepackage{multirow}
\usepackage{algorithm}
\usepackage{algpseudocode}
\usepackage{url}
\PassOptionsToPackage{hyphens}{url}
\usepackage[hidelinks]{hyperref}

\usepackage{mathtools}      

\title{\LARGE \bf
State Estimation Transformers for Agile Legged Locomotion
}

\author{Chen Yu$^{1*}$, Yichu Yang$^{2}$, Tianlin Liu$^{2}$, Yangwei You$^{2}$, Mingliang Zhou$^{2}$, and Diyun Xiang$^{2\dag}$
\thanks{*Work done during the internship at Xiaomi Robotics Lab.}
\thanks{$^{1}$Center for Robotics and Biosystems, Northwestern University, Evanston, IL, USA
}%
\thanks{$^{2}$Xiaomi Robotics Lab, Beijing, China
}%
\thanks{†Corresponding author. Email: xiangdiyun@gmail.com
}%
}

\usepackage[font=small,labelfont=bf,tableposition=top]{caption}
\let\oldtwocolumn\twocolumn
\renewcommand\twocolumn[1][]{%
    \oldtwocolumn[{#1}{
    \begin{center}
           \includegraphics[width=\textwidth]{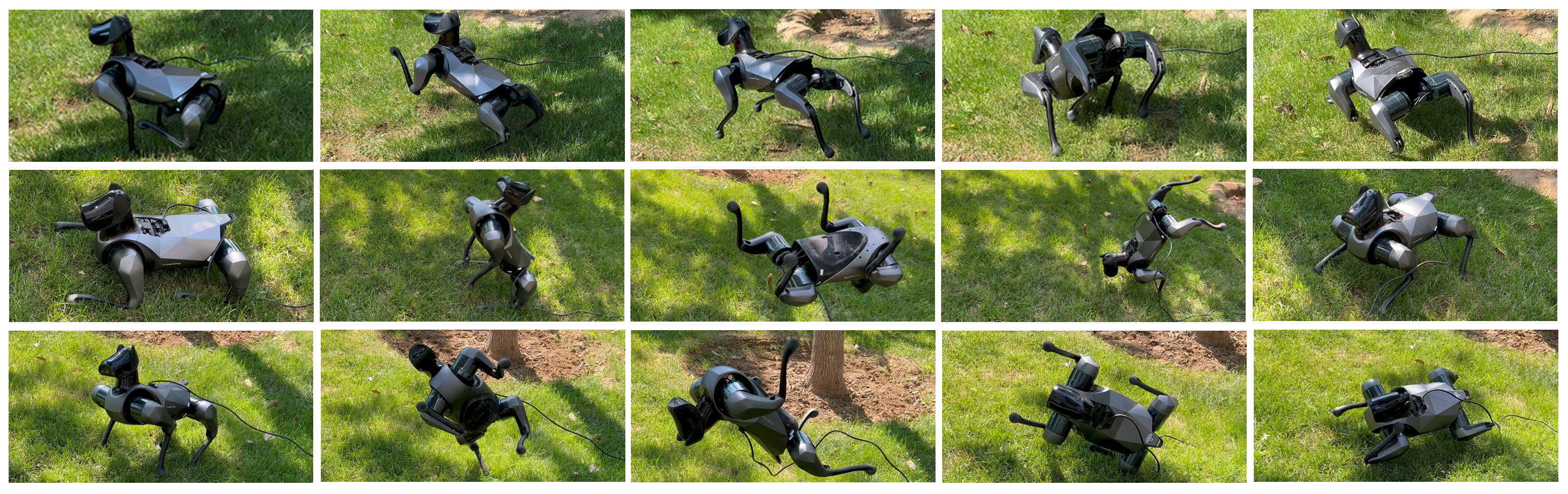}
           \captionof{figure}{We demonstrate the effectiveness of the State Estimation Transformer (SET) in pushing the limit of agile locomotion on several challenging jumping tasks. The Cyberdog2 robot can successfully jump while running (the first row), backflip while running (the second row), and sideflip while running (the third row) in the wild.}
           \label{fig:jump}
        \end{center}
    }]
}

\begin{document}

\maketitle


\thispagestyle{empty}
\pagestyle{empty}


\begin{abstract}
We propose a state estimation method that can accurately predict the robot's privileged states to push the limits of quadruped robots in executing advanced skills such as jumping in the wild.
In particular, we present the State Estimation Transformers (SET), an architecture that casts the state estimation problem as conditional sequence modeling. SET outputs the robot states that are hard to obtain directly in the real world, such as the body height and velocities, by leveraging a causally masked Transformer. By conditioning an autoregressive model on the robot's past states, our SET model can predict these privileged observations accurately even in highly dynamic locomotions.
We evaluate our methods on three tasks --- running jumping, running backflipping, and running sideslipping --- on a low-cost quadruped robot, Cyberdog2.
Results show that SET can outperform other methods in estimation accuracy and transferability in the simulation as well as success rates of jumping and triggering a recovery controller in the real world, suggesting the superiority of such a Transformer-based explicit state estimator in highly dynamic locomotion tasks.

\end{abstract}

\section{Introduction}


Legged animals exhibit remarkable agility --- such as galloping, bounding, and jumping --- inspiring researchers in the field of legged robots to replicate their impressive locomotion skills. Nevertheless, achieving comparable capabilities in robotic systems has proven to be a persistent challenge for the robotics community \cite{smith2023learning}. Reinforcement learning (RL) offers a potent framework for autonomously acquiring skills in robotics.

For quadruped walking control, RL algorithms have demonstrated their effectiveness and robustness in challenging terrains. One of the most popular methods is the student-teacher architecture \cite{lee2020learning}, which trains a teacher policy with privileged information such as height maps and then trains a student policy to reproduce the teacher’s output using non-privileged observations. This architecture and its variants have allowed quadruped robots to traverse multiple challenging terrains, such as stairs \cite{nahrendra2023dreamwaq}, sand~\cite{lai2023sim, choi2023learning}, mattresses~\cite{kumar2021rma}, and curbs \cite{agarwal2023legged}. Another promising branch of RL learns behaviors from motion capture data via imitation learning. For example, Adversarial Motion Priors (AMP) \cite{peng2021amp, escontrela2022adversarial} uses a discriminator to predict the style reward that encourages physically plausible behaviors. These methods have seen many successes in controlling the robots both robustly and naturally \cite{wu2023learning, wang2023amp, wu2023learning1}.

Besides quadruped walking, these RL algorithms have also endowed quadruped robots with more dynamic skills, such as biped stepping \cite{fuchioka2023opt}, bipedal walking \cite{smith2023learning}, jumping \cite{smith2023learning}, backflipping \cite{fuchioka2023opt, li2023learning}, soccer shooting \cite{ji2022hierarchical}, and even extreme parkour \cite{cheng2023extreme, zhuang2023robot, hoeller2023anymal, rudin2022advanced}. These agile behaviors show the effectiveness of RL-based controllers, enabling legged robots to navigate complex environments that are typically accessible to humans.


However, these existing control approaches for quadrupedal locomotion depend on accurate state estimation \cite{ji2022concurrent}, and because such estimations are usually not reliable for dynamic locomotion, they can limit the potential of RL algorithms to explore some more agile behaviors, such as dynamic jumping in the parkour. Ji et al. \cite{ji2022concurrent} explicitly train a state estimator while concurrently training the policy, leading to a more robust running controller even on a slippery plate. Before that, Kim et al. \cite{kim2021legged} define this problem as a Maximum A Posteriori (MAP) estimation problem and solve it with the Gauss-Newton algorithm; Hartley et al. \cite{hartley2020contact} develop a contact-aided invariant extended Kalman filter using the theory of Lie groups and invariant observer design.
Compared with some implicit state estimations such as the student-teacher architecture \cite{lee2020learning, kumar2021rma}, these interpretable state estimators can be used in
conjunction with other modules that also require state information and reduce computational complexity \cite{ji2022concurrent}. However, none of these state methods have shown their ability in reliable estimation in the case of agile skills with aerial phases and unexpected contacts such as jumping. Also, none of these previous works have demonstrated the potential of explicit state estimators in transferring to other tasks.

There are recent works formulating RL as a sequence modeling problem \cite{srivastava2019training,reed2022generalist}. Decision Transformer (DT) \cite{chen2021decision, reid2022can, zheng2022online} uses state, action, and the sum of future rewards as tokens in a Transformer model.
Trajectory Transformer \cite{janner2021offline} uses a Transformer model to learn the dynamics of a robot and uses beam search \cite{reddy1977speech} for planning. Embodiment-aware Transformer \cite{yu2023multi} applies a variant of DT to shape varying robots and Terrain Transformer \cite{lai2023sim} applies a variant of DT to student-teacher architecture for robust locomotion control.
These Transformer-based methods have attained comparable or superior results in standard evaluation tasks when compared to traditional reinforcement learning algorithms, owing to the model's capacity and the self-attention mechanism.

In this work, rather than directly using Transformers to solve the whole control problems, we take advantage of Transformer models to design a state estimator to allow robots to perform more dynamic and agile behaviors without perfect environment information.
In particular, we present the State Estimation Transformer (SET), an architecture that casts the state estimation problem as conditional sequence modeling. SET uses a Transformer architecture to model distributions over robots' non-privileged states and privileged states and outputs the estimated privileged states by leveraging the causally masked Transformer.

We use three jumping tasks to show the advantages of our SET methods compared with traditional MLP estimators; we show the benefits of such an explicit state estimator compared with implicit state estimation by a cross-transferring experiment and additionally deploying a reset policy on the real robots.

The key contributions of this work are: 
\begin{itemize}
 \item Proposing a novel state estimation algorithm, SET;
 \item Designing and training a set of deployable jumping skills leveraging SET;
 \item Evaluating the estimation accuracy and transferability of SET and its alternatives.
\end{itemize}

To the best of our knowledge, this is the first time that Transformers are used for state estimation of a legged robot, and this is the first time that a set of advanced jumping skills including running backflipping and running sideflipping are deployed on a low-cost quadruped robot, as shown in Fig. \ref{fig:jump}.

\section{Preliminaries}
The Transformer model, originally introduced by Vaswani et al. \cite{vaswani2017attention}, is designed to efficiently model sequential data and has demonstrated remarkable performance across a spectrum of tasks, spanning Natural Language Processing \cite{kitaev2018constituency, liu2018generating} and Computer Vision \cite{liu2022video, meinhardt2022trackformer}. Its architecture comprises a series of stacked self-attention layers with residual connections.

In each self-attention layer, the model takes an input sequence of symbol representations $(x_1, \dots, x_n)$ with a context length of $n$ and transforms it into a sequence of continuous representations $\mathbf{z}=\left(z_{1}, \ldots, z_{n}\right)$. This transformation involves linearly mapping each token to a key $k_{i}$, query $q_{i}$, and value $v_{i}$. The self-attention mechanism computes the output for a given token by weighting the values $v_{j}$ based on the normalized dot product between the query $q_{i}$ and the other keys $k_{j}$, as described by the equation below:
\begin{equation}
z_{i}=\sum_{j=1}^{n} \operatorname{softmax}\left(\frac{\left\{\left\langle q_{i}, k_{j^{\prime}}\right\rangle\right\}_{j^{\prime}=1}^{n}}{\sqrt{d_k}}\right)_{j} \cdot v_{j},
\end{equation}
where $d_k$ represents the dimensionality of the queries and keys. 


\section{State Estimation Transformer}

For a deployable highly dynamic locomotion controller, we present a two-stage training framework consisting of training a policy leveraging privileged state information in the simulation and training a state estimator to predict the privileged state information that is not directly deployable on the real robot. 

\subsection{Markov Decision Process and State Estimation}
\label{sec:mdp}
We model the control of the robot as a Markov decision process (MDP), described by the tuple $(\mathcal{S}, \mathcal{A}, P_E, \mathcal{R})$. The MDP tuple consists of states $s \in \mathcal{S}$, actions $a \in \mathcal{A}$, state transition dynamics $P_E\left(\cdot \mid s, a; e\right)$, and a reward function $r=\mathcal{R}(s, a)$. We use $s_{t}, a_{t}$, and $r_{t}$ to denote state, action, and reward at timestep $t$, respectively. We assume that $s_{t}$ consists of privileged observations $o^\prime_t$ and non-privileged observations $o_t$:  $s_t = (o^\prime_t, o_t)$. A state estimator $\mathbb{f}$ can map the history of non-privileged observations to privileged observations:
\begin{equation}
o^\prime_t = \mathbb{f}(o^H_t),
\end{equation}
where $o^H_t$ represents the non-privileged history observation from the timestep $t-H$ to $t$, with a length of $H$; $o^\prime_t$ represents the estimated state at timestep $t$.

\subsection{State Estimation as a Sequence Modeling Problem}

In this work, we cast the state estimation as a sequence modeling problem. We expect a trajectory representation to enable the Transformer to learn meaningful patterns between robot history observations and the current privileged observation, and the Transformer to conditionally
generate the privileged observation based on history observations at test time. Therefore, we define the following trajectory representation which enables autoregressive training and generation:
\begin{equation}
    \tau = (o_1, o^\prime_1, o_2, o^\prime_2,\dots, o_{T}, o^\prime_{T}).
    \label{eq:trajectory}
\end{equation}

\subsection{State Estimation Transformer}

\begin{figure}[t]
  \centering
  \framebox{\parbox{0.47\textwidth}{
  \centering
  \includegraphics[width=0.47\textwidth]{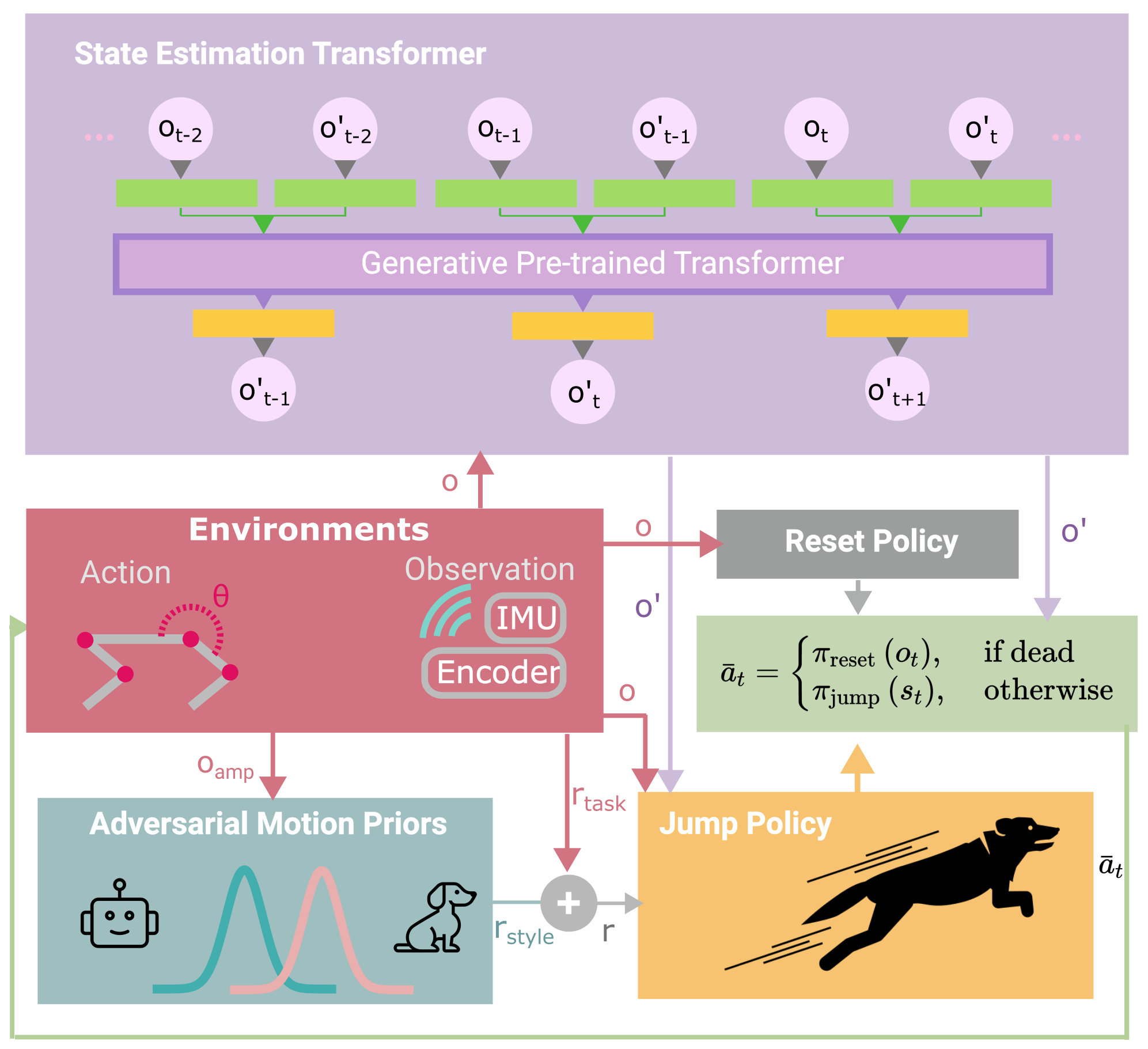}
}}
  \caption{\textbf{State Estimation Transformer architecture in our training pipelines of jumping policies.} First, we train a \textit{Jump Policy} using the task rewards $r_{\text{task}}$ from the \textit{Environments} and the style reward $r_{\text{style}}$ calculated by the \textit{Adversarial Motion Priors}; then we train a \textit{State Estimation Transformer} that leverage a GPT model (green blocks represent embedding and position encoding and yellow blocks represent decoders) to predict the privileged observations. To further exploit the benefit of an accurate explicit state estimator, the predicted privileged observations are used as one of the conditions of triggering a built-in \textit{Reset Policy}.
}
  \label{fig:set}
\end{figure}

\vspace{5pt}
\noindent\textbf{Training.}
The training process of SET is summarized in Alg. \ref{alg:eat}.
We first train the robot using ground true states $s$ in the simulation and use the trained policy $\pi$ to generate the initial trajectory dataset $D$ with the representation in Equation (\ref{eq:trajectory}).
For each training step, we sampled $H$ timesteps of each trajectory from $D$.
In terms of token embeddings, we create linear embedding layers for robot history observation $o$ and privileged observation $o^\prime$. Layer normalization \cite{ba2016layer} is applied in this process and an embedding for each timestep is added to each token.
The tokens, representing these observation pairs, are subsequently fed into a Generative Pre-trained Transformer (GPT) model \cite{radford2018improving}. This model employs an autoregressive approach to predict future token pairs, specifically $(o, o^\prime)$.
The predicted privileged observation $o^\prime$ is instrumental in calculating the mean-squared error during the backpropagation process.




\vspace{5pt}
\noindent\textbf{Evaluation.}

For evaluating the privileged observations while deploying the policy on a real robot,
the robot's initial non-privileged observation will serve as the conditioning information for the initiation of the generation process. This entails supplying SET with the most recent $H$ time steps from the ongoing trajectory denoted as $\tau$ in order to generate a prediction for the privileged observation for the last timestep.

\begin{algorithm}[t]
\caption{State Estimation Transformer (SET)}
\begin{algorithmic}[1]
\Require Initial $D$, trained policy $\pi$
\Ensure Trained SET Model
\For{$i = 0, \dots, I - 1 $ iterations}
\State Sample $H$-long ($o, o^\prime, t$) from $D$
\State Stack embeddings of ($o, o^\prime$) for each timestep
\State Feed the Stacks to the GPT model
\State Update the GPT model
\EndFor
\end{algorithmic}
\label{alg:eat}
\end{algorithm}

\section{Learning To Jump with SET}

SET is a state estimation algorithm for learning complex locomotion skills. 
In this section, we focus on robot jumping and its variants as concrete applications for SET and introduce how our framework can be applied to solve them, as summarized in Fig. \ref{fig:set}.

\subsection{Robot Setup}
We use the Cyberdog2 quadruped robot from Xiaomi \cite{Cyberdog2} as our experimental platform and build our simulation using Isaac Gym \cite{makoviychuk2021isaac}.
The compact size (0.16 m body length) and relatively lightweight (8.9 kg) of the Cyberdog2 robot enable us to tackle difficult dynamic tasks, while its low-cost actuators with a maximum torque limit of 12 Nm also pose challenges of highly agile locomotion control.


\subsection{Learning to Jump}
\vspace{5pt}
Here, we describe our implementation of the first stage of our two-stage training framework.
We want to train a policy to allow the robot to walk/run at desired velocities and jump at command. Similar to previous work, the action $a$ is the desired joint position of the motors, sent to a PD controller. The state $s$ and the reward $r$ are inspired by \cite{InspirAI}, as described below.

\noindent\textbf{Observations.}
As introduced in Section \ref{sec:mdp}, the state $s$ consists of privileged observation $o^\prime$ and non-privileged observation $o$. The privileged observation is simply defined as:
\begin{equation}
    o^\prime = (h, \mathbf{v}),
\end{equation}
where $h$ is the height of the robot's Center of Mass (CoM) and $\mathbf{v}$ is the three-dimensional velocities of the robot. The non-privileged observation is defined as:
\begin{equation}
    o = (\mathbf{\omega}, \mathbf{\phi}, \mathbf{q}, \dot{\mathbf{q}}, \mathbf{p}, \mathbf{cmd}),
\end{equation}
where $\mathbf{\omega}$ and $\mathbf{\phi}$ are the angular velocity and the base orientation; $\mathbf{q}$ and $\dot{\mathbf{q}}$ are the joint positions and velocities; $\mathbf{p}$ is the
Cartesian positions of the feet relative to the robot CoM; and $\mathbf{cmd}$ is a 5-dimensional vector that consists of the velocity commands for the x/y/yaw directions, the target height of jumping, and a Boolean value representing the jump signal. 
The goal of the robot is to track the commanded velocities, and jump to the target height when the jump signal is equal to 1. The jump signal will automatically reset to 0 when the robot reaches the target height. 

The observations for the AMP discriminator $o_{\text{AMP}}$ is the full state $s$ except for the command vector:
\begin{equation}
    o_{\text{AMP}} = (\mathbf{\omega}, \mathbf{\phi}, \mathbf{q}, \dot{\mathbf{q}}, \mathbf{p}).
\end{equation}
Data of such observations from the robots and from the motion capture dataset of a jumping and galloping Shepherd dog \cite{zhang2018mode} will be used to train the AMP discriminator.

\begin{table}[t]
\caption{Reward structure.}
\label{tab:reward}
\begin{tabular}{|c|c|ccc|}
\hline
\multirow{2}{*}{Reward} & \multirow{2}{*}{Expression} & \multicolumn{3}{c|}{Weight} \\ \cline{3-5} 
 &  & \multicolumn{1}{c|}{RJ} & \multicolumn{1}{c|}{RB} & RS \\ \hline
Linear Velocity & 
\scalebox{0.6}{$\phi\left(\mathbf{v}_{b, x y}^*-\mathbf{v}_{b, x y}\right)$}
& \multicolumn{1}{c|}{20} & \multicolumn{1}{c|}{20} & 20 \\ \hline
Angular Velocity & 
\scalebox{0.6}{$\phi\left(\boldsymbol{\omega}_{b, z}^*-\boldsymbol{\omega}_{b, z}\right)$}
& \multicolumn{1}{c|}{6.66} & \multicolumn{1}{c|}{6.66} & 10 \\ \hline
Jump Height & 
\scalebox{0.6}{$
\begin{aligned}
    \begin{dcases*}
    \phi(h_{jump}^\star-h) \cdot \text{clip}(v_x), & if $ sig = 1 $,\\
    \phi(h_{walk}^\star-h), & otherwise,
    \end{dcases*}
\end{aligned} 
$ }
& \multicolumn{1}{c|}{5} & \multicolumn{1}{c|}{5} & 5 \\ \hline
Jump Goal & 
\scalebox{0.6}{$
\begin{aligned}\label{eqn:goal}
    \begin{dcases*}
    \text{clip}(v_x), & if $ |h^\star-h|<\epsilon_h$ and $sig = 1$,\\
    0, & otherwise. 
    \end{dcases*}
\end{aligned}
$}
& \multicolumn{1}{c|}{100} & \multicolumn{1}{c|}{100} & 100 \\ \hline
Jump Height (Roll) & 
\scalebox{0.6}{$
\begin{aligned}
    \begin{dcases*}
    \phi(h_{jump}^\star-h) \cdot |\phi_{\text{roll}}|, & if $ sig = 1 $,\\
    0, & otherwise. 
    \end{dcases*}
  \end{aligned}
$}
& \multicolumn{1}{c|}{0} & \multicolumn{1}{c|}{0} & 5 \\ \hline
Jump Goal (Roll) & 
\scalebox{0.6}{$
\begin{aligned}
    \begin{dcases*}
    |\phi_{\text{roll}}|, & if $ |h^\star-h|<\epsilon_h$ and $sig = 1 $,\\
    0, & otherwise. 
    \end{dcases*}
\end{aligned}
$}
& \multicolumn{1}{c|}{0} & \multicolumn{1}{c|}{0} & 1 \\ \hline
Pitch Reward & 
\scalebox{0.6}{$
\begin{aligned}
    \begin{dcases*}
    \phi_{\text{pitch}}, & if $ falling = 1 $,\\
    0, & otherwise. 
    \end{dcases*}
\end{aligned}
$}
& \multicolumn{1}{c|}{10} & \multicolumn{1}{c|}{50} & 0 \\ \hline
Jump Forward & 
\scalebox{0.6}{$
\left(\operatorname{angle}\left(\phi_{yaw,jumping}, \phi_{yaw,landing}\right)\right)^2
$}

& \multicolumn{1}{c|}{0} & \multicolumn{1}{c|}{0} & -400 \\ \hline
Feet Air Time & 
\scalebox{0.6}{$\sum_{f=0}^4\left(\mathbf{t}_{a i r, f}-0.5\right)$}
& \multicolumn{1}{c|}{5} & \multicolumn{1}{c|}{5} & 50 \\ \hline
Action Rate & 
\scalebox{0.6}{$-\left\|\mathbf{q}_j^*\right\|^2$}
& \multicolumn{1}{c|}{0} & \multicolumn{1}{c|}{-10} & -10 \\ \hline
\end{tabular}
\end{table}

\vspace{5pt}
\noindent\textbf{Rewards.}
Besides the common tracking terms similar to previous works in locomotion learning \cite{rudin2022learning, ji2022concurrent, kumar2021rma} and a style reward evaluated by the AMP discriminator~\cite{escontrela2022adversarial, wu2023learning}, we define different reward functions for three different expected jumping gaits that are inspired by human Parkour:

\begin{itemize}
    \item \textit{Running Jump (RJ): }The robot jumps forward while maintaining its running momentum.
    \item \textit{Running Backflip (RB): }The robot performs a backflip while running forward and seamlessly continues forward motion upon completing the backflip.
    \item \textit{Running Sideflip (RS): }The robot executes a side-flip while in forward motion and smoothly resumes running forward after completing the side-flip.
\end{itemize}

For the \textit{Running Jump} gait, we expect the robot to jump as close as the target height $h_{jump}^\star$ when the jump signal $ jump\_sig = 1 $ and keep close to the walking target height $h_{walk}^\star$ when $ jump\_sig = 0 $. Therefore, we design the jumping reward as follows:
\begin{align}\label{eqn:jump}
    \begin{dcases*}
    \phi(h_{jump}^\star-h) \cdot \text{clip}(v_x,-0.5,2), & if $ jump\_sig = 1 $,\\
    \phi(h_{walk}^\star-h), & otherwise,
    \end{dcases*}
\end{align}
where $\phi(x):=\exp \left(-\frac{\|x\|^2}{0.25}\right)$ and $v_x$ represents the forward velocity.
In addition, to encourage the robot to reach the jumping goal, we add a bonus to the robot while reaching the goal:
\begin{align}\label{eqn:bonus}
    \begin{dcases*}
    \text{clip}(v_x,-0.5,2), & if $ |h_{jump}^\star-h|<\epsilon_h$ and $jump\_sig = 1 $,\\
    0, & otherwise. 
    \end{dcases*}
\end{align}
As the robot has only limited torques, we additionally design a pitch reward to prevent the robot from cheating by standing on two legs as follows in the falling phase of the jumping:
\begin{align}\label{eqn:pitch}
    \begin{dcases*}
    \phi_{\text{pitch}}, & if $ falling = 1 $,\\
    0, & otherwise. 
    \end{dcases*}
\end{align}
The falling phase $ falling = 1 $ is defined as the period between the robot reaching the jumping goal and one of its feet contacts with the ground.

We define the pitch angle $\phi_{\text{pitch}}$ as follows: It varies from 0 radians (normal pose) to $\pi/2$ radians (upward tilt) and then abruptly switches to $-\pi/2$ radians, with a gradual return to 0 radians when rotating backward. 
The beauty of such a pitch reward and definition is that we can use it to switch between the \textit{Running Jump} gait and the \textit{Running backflip} gait: a small weight for this pitch reward can lead to a \textit{Running Jump} gait while a big weight can encourage the robot to backflip when receiving the jump signal.

We also design a variant of the Equation (\ref{eqn:jump}) for the side-flipping task while encouraging the rotation of the robot in the roll angle:
\begin{align}\label{eqn:sidejump}
    \begin{dcases*}
    \phi(h_{jump}^\star-h) \cdot |\phi_{\text{roll}}|, & if $ jump\_sig = 1 $,\\
    0, & otherwise. 
    \end{dcases*}
  \end{align}


Similarly, a successful jump reward for side-flipping is designed as follows:
\begin{align}\label{eqn:sidegoal}
    \begin{dcases*}
    |\phi_{\text{roll}}|, & if $ |h_{jump}^\star-h|<\epsilon_h$ and $jump\_sig = 1 $,\\
    0, & otherwise. 
    \end{dcases*}
\end{align}


Other auxiliary rewards \cite{margolis2023walk} include the feet-in-air reward, action rate penalty, and reward for robot jumping forward in the sideflipping gait are summarized in Tab. \ref{tab:reward}, as well as the weights of all reward terms.




\noindent\textbf{Curriculum in torque limit.}
While we focus on a low-cost quadruped robot with motors with a maximum torque of only 12 Nm, we find that starting from a more relaxed torque limit in the simulation can significantly help the agents' exploration. Therefore, we set the torque limits of all motors as 300 Nm at the beginning of the training and gradually decrease the maximum torque limit to 12 Nm.

\begin{figure*}[t]
  \centering
  \framebox{\parbox{\textwidth}{
  \centering
  \includegraphics[width=\textwidth]{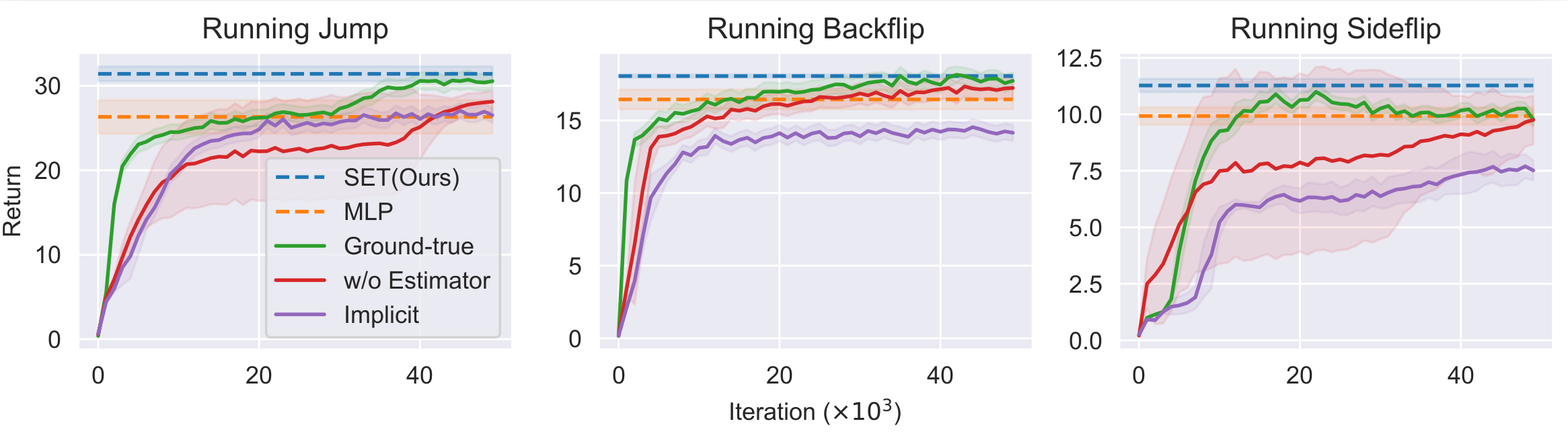}
}}
  \caption{\textbf{Training curves of different algorithms on three jumping gaits.} We compare the training curves using the ground-true full observations (\textit{Ground-true}), only non-privileged observations (\textit{w/o Estimator}), 20 historical stacked non-privileged observations (\textit{Implicit}), and the results using our SET and MLP estimators. Results show that for highly dynamic locomotion, the privileged observations such as robot velocity and height can significantly increase the efficiency and robustness of policy training; and our deployable method can reproduce the results trained with ground-true privileged observations.
}
  \label{fig:train}
\end{figure*}

\subsection{Reset Policy}
\label{sec:reset}
One of the benefits of an explicit state estimator such as SET is that it can be used in conjunction with other modules \cite{ji2022concurrent} although hardly previous works have exploited this feature. In this work, the state estimator can share the estimated results between the jumping policy $\pi_{\text{jump}}$ with another built-in reset policy $\pi_{\text{reset}}$, which can help to reset the robot when it fails the jumping and falls on the ground when the environment is too noisy.

Traditionally, a recovery controller can kick in when the robot is detected as upside-down \cite{katz2019mini, smith2022legged}. However, for our \textit{Jumping Backflip} and \textit{Jumping Sideflip} gaits, additional height information should be used to tell the difference between the normal in-the-air phase or unexpected falling on the ground. Therefore, we have the final actions $\bar{a}$ as follows:
\begin{align}
    \bar{a}_t= 
    \begin{dcases*}
    \pi_{\text{reset}}(o_t), & if $ \mathbf{g}^p \cdot \mathbf{g}<\epsilon_r$ and $h < h_{\text{walk}}^\star$,\\
    \pi_{\text{jump}}(s_t), & otherwise,
    \end{dcases*}
\end{align}
\begin{align}
    \bar{a}_t= 
    \begin{dcases*}
    \pi_{\text{reset}}(o_t), & if $dead$,\\
    \pi_{\text{jump}}(s_t), & otherwise
    \end{dcases*}
\end{align}
where $\mathbf{g}^p \cdot \mathbf{g}$ is the cosine similarity between the projected gravity $\mathbf{g}^p$ and the gravity vector (0, 0, -1) $\mathbf{g}$.

\section{Simulation Results}

\begin{table}[tb]
\caption{Tracking and estimation error for a walking policy}
\label{tab:walk}
\resizebox{\columnwidth}{!}{
\begin{tabular}{|l|l|l|l|}
\hline
Task                               & Model         & $V_X$ [m/s] & $V_Y$ [m/s] \\ \hline
\multirow{5}{*}{Tracking Error} & SET (Ours)    &  0.0606  &  0.0676  \\ \cline{2-4} 
                                   & MLP           &  0.1621  &  0.1636  \\ \cline{2-4} 
                                   & w/o Estimator &  2.620  &  1.315  \\ \cline{2-4} 
                                   & Implicit      &  1.811  &  0.9966  \\ \cline{2-4} 
                                   & Ground-true   &  0.0598  &  0.0610  \\ \hline
\multirow{4}{*}{Estimation Error}  & SET (Ours)    &  0.0179  &  0.0172  \\ \cline{2-4} 
                                   & MLP           &  0.1663  &  0.1445  \\ \cline{2-4} 
                                   & SET-1e4        &  0.1323  &  0.06822  \\ \cline{2-4} 
                                   & SET-H1        &  0.08784  &  0.07742  \\ \cline{2-4} 
                                   \hline
\end{tabular}
}
\end{table}

\begin{table*}[bt]
\caption{Estimated errors of different jumping gaits}
\label{tab:errors}
\centering
\resizebox{\textwidth}{!}{%
\begin{tabular}{|l|c|c|c|c|c|c|c|c|c|c|c|c|}
\hline
\multirow{2}{*}{} & \multicolumn{4}{c|}{Jumping} & \multicolumn{4}{c|}{Backflipping} & \multicolumn{4}{c|}{Rolling} \\ \cline{2-13} 
 & \multicolumn{1}{c|}{$h$} & \multicolumn{1}{c|}{$v_x$} & \multicolumn{1}{c|}{$v_y$} & $v_z$ & \multicolumn{1}{c|}{$h$} & \multicolumn{1}{c|}{$v_x$} & \multicolumn{1}{c|}{$v_y$} & $v_z$ & \multicolumn{1}{c|}{$h$} & \multicolumn{1}{c|}{$v_x$} & \multicolumn{1}{c|}{$v_y$} & $v_z$ \\ \hline
SET (Ours) & \textbf{0.001091} & \textbf{0.01393} & \textbf{0.0065} & \textbf{0.0091} & \textbf{0.001438} & \textbf{0.01873} & \textbf{0.007415} & \textbf{0.01218} & \textbf{0.001113} & \textbf{0.00905} & \textbf{0.007131} & \textbf{0.00686} \\ \hline
MLP & 0.007877 & 0.09994 & 0.04037 & 0.06498 & 0.003593 & 0.05691 & 0.02224 & 0.03335 & 0.005348 & 0.04553 & 0.03514 & 0.03766 \\ \hline
\end{tabular}%
}
\end{table*}

\subsection{Walking}
\label{sec:walk}
Firstly, we evaluate our estimator in a simple walking scenario with a similar experimental setup as in \cite{ji2022concurrent}. We train a walking policy using PPO on a flat plane. The observation is the same as our jumping observation with $jump\_sig=0$ and the reward function only contains the velocity tracking terms. We train the robot with full observation $s$ and then train a state estimator based on the trajectories of the last trained policy.
Such a setting is called a \textit{sequential model} in~\cite{ji2022concurrent}. The number of historical steps $H$ we used for the SET is 20 and for MLP is 5 since the performance of MLP will drop significantly when $H$ increases; we set the number of blocks as 6 for SET and the number of layers as 3 for MLP since this is also the best parameters we have tuned for MLP.
We also evaluate the performance of the following methods:
\begin{itemize}
    \item \textit{Ground-true}: we train a policy with the ground-truth privileged robot states and non-privileged states;
    \item \textit{w/o Estimator}: we train a policy with only non-privileged states;
    \item \textit{Implicit}: we train a policy with 20 stacked historical non-privileged states;
    \item \textit{SET}: we train a SET based on trajectories of the final policy using ground-true privileged observations;
    \item \textit{MLP}: we train an MLP estimator based on trajectories of the final policy using ground-true privileged observations.
\end{itemize}

Tab. \ref{tab:walk} shows the tracking and estimation error for the walking policy. SET demonstrates a lower command following errors and estimation errors compared to MLP, \textit{w/o Estimator}, and \textit{Implicit}, which is similar to the \textit{Ground-true}.

We hypothesize that the superiority of SET comes from the context information of previous tokens and the capacity of the model for fitting our diverse training dataset that contains both varying trajectories.
To investigate the importance of access to previous non-privileged states, we ablate on the context length $H$. 
The performance of SET degrades significantly when the number of historical observations $H$ is 1 (SET-H1 in Tab. \ref{tab:walk}), indicating that past information is essential for this state estimation task.

We further investigate the impact of the training dataset on the performance of SET. We decrease the number of trajectories in the training dataset from $1\times10^5$ to $1\times10^4$, denoted as Model \textit{SET-1e4} in Tab. \ref{tab:walk}. This reduction of dataset size also yields degraded performance of SET, suggesting that the high complicity of SET allows it to leverage all the information in the dataset and implicitly build the associations between non-privileged observations and privileged observations via the similarity of the query and key vectors. This makes it superior in privileged state estimation. 

\begin{figure*}[t]
  \centering
  \framebox{\parbox{\textwidth}{
  \centering
  \includegraphics[width=\textwidth]{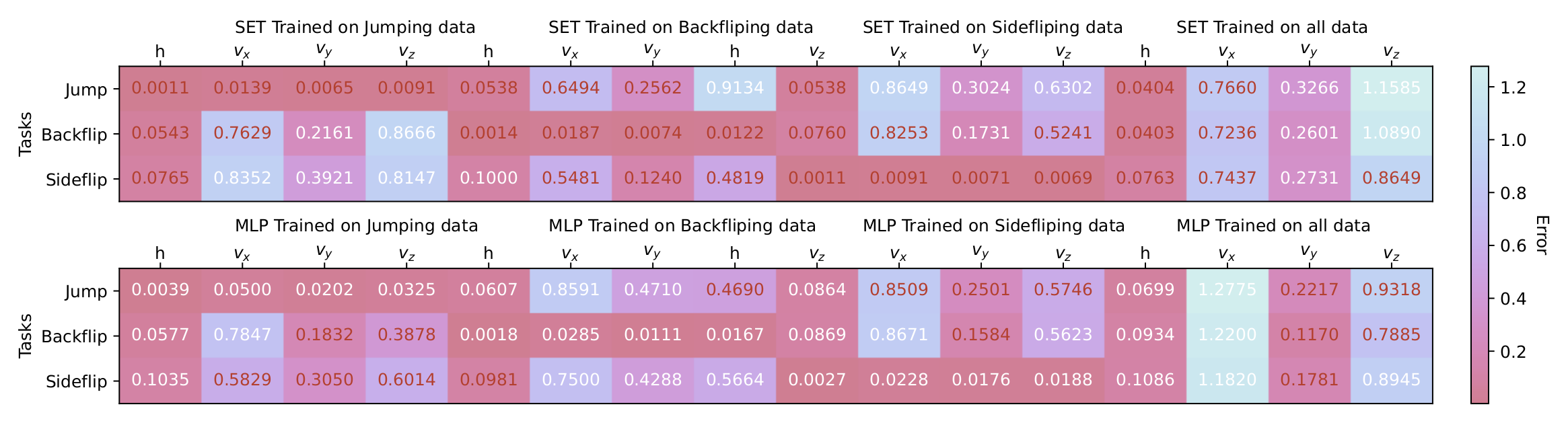}
}}
  \caption{\textbf{Prediction error of the estimators that are trained on different data and applied on different tasks.} Each pair of training dataset / application task correspond to four adjacent items in the heatmaps, representing errors in the estimation of $h$, $v_x$, $v_y$, and $v_z$. We show the results from SET and MLP to compare their abilities in generalization, while SET outperforms MLP in 2/3 (32/48, highlighted in red) cases. These results suggest the abilities of a trained explicit estimator to be applied to another task; and demonstrate the superiority of SET over MLP in explicit state estimation.
}
  \label{fig:cross}
\end{figure*}

\subsection{Jumping}
\label{sec:train}
\vspace{5pt}
\noindent\textbf{Learning to Jump.} To evaluate and analyze the proposed methods, we test the performance of SET and alternative methods \textit{w/o Estimator}, \textit{Implicit}, and \textit{MLP} as well as \textit{Ground-true} as described in Section \ref{sec:walk} for training three jumping gaits.

Fig. \ref{fig:train} shows the training curves of different algorithms on three jumping gaits, averaged over three trials. Results show that for all of these three jumping gaits, the privileged observations are essential: agents with the explicit ground-true privileged observations show a much more efficient and stable training process compared with using stacked historical non-privileged observation (\textit{Implicit}) or without privileged observations (\textit{w/o Estimator}). The results of SET also show that SET can reproduce the performance of the policy trained with privileged observations, even if it only requires non-privileged observations. These results are consistent with that in Section \ref{sec:walk}.

\vspace{5pt}
\noindent\textbf{State Estimation.}
We further compare the estimated errors between SET and MLP estimators on these three jumping tasks, as shown in Tab. \ref{tab:errors}. We demonstrate the RMS errors averaged over 2048 trials on three jumping tasks and four dimensions of predicted privileged observations. Results show that SET can significantly outperform MLP estimators in all of the cases.

\subsection{Transferring}

As we have mentioned in Section \ref{sec:reset}, one of the advantages of explicit estimators is that their estimated results can be used for different downstream tasks; this feature can be further highlighted by the transferability of Transformers. To test the transferability of different estimators, we evaluate the performance of estimators on the training tasks that may be different from the training dataset.

In Fig. \ref{fig:cross}, we show the estimated errors of implementing the estimator that are trained on different training datasets (x-axis) on different tasks (y-axis). Results show that in 2/3 cases SET can predict the privileged observations better than MLP. Interestingly, for estimators that all trained on all data, SET shows a significantly more accurate estimation in body height and x velocity --- which are the two most important privileged information for jumping tasks --- than MLP, making it more suitable than MLP to train a distilled estimator that can be used for all tasks for such agile and dynamic tasks.

\section{Experimental Results}
We directly deploy the policy trained with privileged observations on the real robot and use SET or MLP estimator to estimate those privileged states.
Note that although alternative methods such as the \textit{Stacked obs} evaluated in Section \ref{sec:train} also do not require privileged observations for their policy, it is still challenging to deploy them on the real robots since we need the height information to reset the jump signal.

\subsection{Jumping in the Real-world}
All three jumping policies with SET can be successfully deployed on the real robots as shown in Fig. \ref{fig:jump}: Our robots can robustly jump, backflip, and sideflip seamlessly during a dynamic running process in the wild as commanded. More demonstrations can be seen in the supplementary videos.

\begin{table}[b]
\caption{Success rates in the real world}
\label{tab:real}
\centering
\resizebox{\columnwidth}{!}{%
\begin{tabular}{|c|c|c|c|c|c|}
\hline
 & Jump & Backflip & Sideflip & Distillation & Reset \\ \hline
SET & 75.0\% & \textbf{85.0\%} & \textbf{85.0\%} & \textbf{85.0\%} & \textbf{100.0\%} \\ \hline
MLP & \textbf{85.0\%} & 80.0\% & 70.0\% & 60\% & 75.0\% \\ \hline
\end{tabular}%
}
\end{table}

Quantitatively, we compare the success rates of jumping using SET and MLP estimators. Each jumping policy is tested twenty times, and the overall success rate is shown in Tab. \ref{tab:real}. We show the success rates of the \textit{Running Jump} policy, \textit{Running Backflip} policy, and \textit{Running Sideflip} policy paired with SET or MLP estimators on the real robots, as well as the overall success rates using the distilled policy trained with all jumping data and evaluating it on all jumping tasks.

Results show that SET can outperform MLP on the Backfliping and Sidefliping tasks when the policy is trained using a task-specific dataset; SET can also significantly outperform MLP when a general-purpose estimator is trained using training data from all tasks, which suggests the superiority of SET in convenient real-robot deployment. For the jumping gait, the lower success rate observed in SET may be attributed to its longer observation horizon, which makes it more sensitive to real-world noise. 

\subsection{Reseting at Failure}
As described in Section \ref{sec:reset}, we use the predicted robot height from the estimators as one of the conditions for triggering the reset policy.
Note that since our policy can have an almost 100\% success rate in the simulators, our estimators have limited training samples of failed jumpings in the training phase. Therefore, this is also a test of the generalization ability and sample efficiency of the estimators.

\begin{figure}[t]
  \centering
  \framebox{\parbox{0.47\textwidth}{
  \centering
  \includegraphics[width=0.47\textwidth]{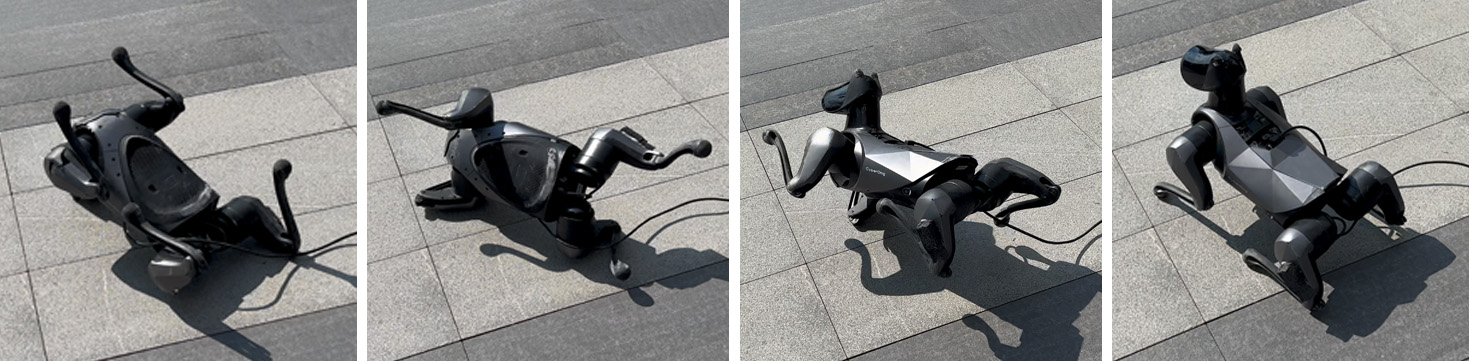}
}}
  \caption{\textbf{Recovering from a failed jumping.} The estimated robot height from the estimator is used as one of the conditions for triggering the reset policy.
}
  \label{fig:reset}
\end{figure}

Fig. \ref{fig:reset} shows the snapshots of a robot resetting from a failed jumping. We also compare the successful rate of resetting between using SET and MLP as the height estimator in Tab. \ref{tab:real}. SET also outperforms MLP in this application in real-world deployment, which is consistent with the results of estimation accuracy (Tab. \ref{tab:errors}) and generalization ability (Fig. \ref{fig:cross}) in simulation. 

\section{Conclusions}

In this work, we propose a state estimation method SET for agile locomotion control. We pose the state estimation challenge as a conditional sequence modeling problem, and use Transformer to fit a data sequence of (non-privileged observations, privileged observations) tuples.
The proposed methods are evaluated on three AMP-based jumping tasks in a two-stage training manner: Firstly the policies are trained with privileged observations and then estimators are trained to predict these observations.
After showing that privileged observations are essential information for such dynamic and agile locomotion tasks, we demonstrate that SET can show superiority in estimation accuracy, transferability, and real-world success rates compared to MLP.
To further exploit the benefits of explicit state estimation, we use the output from the estimator as one of the conditions of performing a recovery controller, showing the practical application of our estimators in real-world applications.
It would be our future work to integrate these agile locomotion skills into an autonomous planning framework to unlock the full potential for real-world application of low-cost quadruped robots.

\addtolength{\textheight}{-9cm}   


\bibliographystyle{IEEEtran}
\bibliography{IEEEabrv,ref}







\end{document}